\documentclass{article} 
\usepackage{iclr2026_conference,times}

\usepackage{amsmath,amsfonts,bm}

\def\eqref#1{equation~\ref{#1}}

\def\1{\bm{1}}

\DeclareMathAlphabet{\mathsfit}{\encodingdefault}{\sfdefault}{m}{sl}
\SetMathAlphabet{\mathsfit}{bold}{\encodingdefault}{\sfdefault}{bx}{n}

\usepackage{natbib}
\PassOptionsToPackage{table}{xcolor}
\usepackage{hyperref}
\usepackage{url}
\usepackage{booktabs}
\usepackage{graphicx}
\usepackage[ruled,vlined]{algorithm2e}
\usepackage{algorithmic}
\usepackage{amsmath} 
\usepackage{amsthm} 
\usepackage{tikz}
\usepackage{multirow}
\usepackage{xcolor}
\usepackage{colortbl}
\usepackage{amssymb}
\usepackage{pifont} 
\usepackage{makecell}
\usepackage[normalem]{ulem}
\usepackage{fancyhdr}

\renewcommand{\headrulewidth}{0pt}  
\renewcommand{\footrulewidth}{0pt}  

\title{\our: Beyond Sparsity in Diffusion Transformers via Fine-Tunable Sparse–Linear Attention}

\author{
\hspace{-.2em}Jintao Zhang, Haoxu Wang, Kai Jiang, Shuo Yang, Kaiwen Zheng, Haocheng Xi, Ziteng \\
\textbf{Wang, Hongzhou Zhu, Min Zhao, Ion Stoica, Joseph E. Gonzalez, Jun Zhu, Jianfei Chen} \\
Tsinghua University, UC Berkeley \\
\texttt{\{zhang-jt24@mails., jianfeic@, dcszj@\}tsinghua.edu.cn}
}

\newcommand{\vQ}{\mathbf{Q}}
\newcommand{\vK}{\mathbf{K}}
\newcommand{\vV}{\mathbf{V}}
\newcommand{\vdQ}{\mathbf{dQ}}
\newcommand{\vdK}{\mathbf{dK}}
\newcommand{\vdV}{\mathbf{dV}}
\newcommand{\vS}{\mathbf{S}}
\newcommand{\vdS}{\mathbf{dS}}
\newcommand{\vP}{\mathbf{P}}
\newcommand{\vdP}{\mathbf{dP}}

\newcommand{\vO}{\mathbf{O}}
\newcommand{\vdO}{\mathbf{dO}}
\newcommand{\vM}{\mathbf{M}}
\newcommand{\vH}{\mathbf{H}}
\newcommand{\vZ}{\mathbf{Z}}
\newcommand{\vdH}{\mathbf{dH}}
\newcommand{\vdZ}{\mathbf{dZ}}
\newcommand{\vD}{\mathbf{D}}

\newcommand{\annotate}[1]{\textcolor{gray}{{#1}}\xspace}

\newcommand{\our}{\texttt{SLA}\xspace}

\definecolor{deepgreen}{rgb}{0.0, 0.5, 0.0}  
\definecolor{deepred}{rgb}{0.6, 0.0, 0.0}  
\definecolor{darkgreen}{rgb}{0.15, 0.75, 0.15}
\definecolor{cvprblue}{rgb}{0.21,0.49,0.74}
\definecolor{lightblue}{rgb}{0.90, 0.95, 0.99}

\NewDocumentCommand{\jintao}{ mO{} }{\textcolor{blue}{\textsuperscript{\textit{JT}}\textsf{\small[#1]}}}

\NewDocumentCommand{\ziteng}{ mO{} }{\textcolor{orange}{\textsuperscript{\textit{ZT}}\textsf{\small[#1]}}}

\NewDocumentCommand{\jianfei}{ mO{} }{\textcolor{orange}{\textsuperscript{\textit{JF}}\textsf{\small[#1]}}}

\NewDocumentCommand{\haocheng}{ mO{} }{\textcolor{orange}{\textsuperscript{\textit{HC}}\textsf{\small[#1]}}}

\NewDocumentCommand{\haoxu}{ mO{} }{\textcolor{blue}{\textsuperscript{\textit{HX}}\textsf{\small[#1]}}}

\iclrfinalcopy 
\begin{document}

\maketitle

\begin{abstract}
In Diffusion Transformer (DiT) models, particularly for video generation, attention latency is a major bottleneck due to the long sequence length and the quadratic complexity. 
Interestingly, we find that attention weights can be decoupled into two matrices: a small fraction of large weights with high rank and the remaining weights with very low rank. This naturally suggests applying sparse acceleration to the first part and low-rank acceleration to the second.
Based on this finding, we propose \our (\textbf{S}parse-\textbf{L}inear \textbf{A}ttention), a trainable attention method that fuses sparse and linear attention to accelerate diffusion models. \our classifies attention weights into critical, marginal, and negligible, applying $\mathcal{O}(N^2)$ attention to critical weights, $\mathcal{O}(N)$ attention to marginal weights, and skipping negligible ones. \our combines these computations into a single GPU kernel and supports both forward and backward passes. With only a few fine-tuning steps using \our, DiT models achieve a $\bf 20\times$ reduction in attention computation, resulting in significant acceleration without loss of generation quality. Experiments show that \our reduces attention computation by $\bf 95\%$ without degrading end-to-end generation quality, outperforming baseline methods. In addition, we implement an efficient GPU kernel for \our, which yields a $\bf 13.7\times$ speedup in attention computation and a $\bf 2.2\times$ end-to-end speedup in video generation on Wan2.1-1.3B. The code will be available at \url{https://github.com/thu-ml/SLA}.
\end{abstract}

\section{Introduction} \vspace{-.5em}
Among the operations in Transformers, attention~\citep{vaswani2017attention} is the only one with quadratic computation complexity, while others mostly scale linearly with the sequence length $N$. In Diffusion Transformer (DiT) models~\citep{Peebles2022DiT}, especially for video generation, attention becomes the primary computational bottleneck, as the sequence length typically ranges from 10K to 100K. Reducing the cost of attention is therefore critical for improving the efficiency of DiT models. Existing efficient attention methods~\citep{zhangsurvey} for DiTs fall into two main categories: (1) numerous \emph{sparse attention} methods~\citep{li2025radial,zhang2025spargeattn,xi2025sparse,yang2025sparse,zhang2025vsa,wu2025vmoba,shen2025draftattention,hassani2023neighborhood,liu2025fpsattention}, which compute only a subset of attention scores, and (2) a few \emph{linear attention} methods~\citep{xie2024sana,zhu2025dig}, which reformulate the operation to achieve $\mathcal{O}(N)$ complexity.

\textbf{Limitation.} Despite recent progress, both approaches face challenges in substantially reducing attention computation:  
\textbf{(L1)} Linear attention methods often fail in practice, especially on video diffusion models. Existing work on linear attention in diffusion is rare and primarily limited to image generation. Our experiments show that when applied to diffusion models, particularly video generation, linear attention severely degrades video quality. 
\textbf{(L2)} Sparse attention methods rarely achieve very high sparsity and require a considerable fraction of the full complexity of attention. In practice, they typically reach only 40--60\% sparsity for sequence length below 50K. Although some recent works~\citep{yang2025sparse,li2025radial} report sparsity of 80--85\%, such results are obtained on very long sequences (e.g., 100K--300K), where achieving high sparsity is easier.  

\textbf{Key Observation.} We find that attention weights in diffusion transformers can be decomposed into two matrices: a small fraction of large weights with high rank and a large fraction of the remaining weights with extremely low rank. This explains why sparse attention or linear attention alone cannot achieve satisfactory results and naturally suggests applying sparse acceleration to the first part and low-rank acceleration to the second.

\textbf{Our Method.} Based on the observation above, we propose \our, a trainable hybrid sparse and linear attention for DiT models. 
Specifically, attention weights are partitioned into blocks and dynamically classified into three categories: critical, marginal, and negligible. Critical blocks are computed exactly using FlashAttention, negligible blocks are skipped, and, unlike existing methods, marginal blocks are processed with linear attention. This design allows sparsity to increase dramatically (e.g., 70\%$\rightarrow$95\%) while maintaining accuracy. Since linear attention is computationally negligible, costing less than 0.5\% of full attention in video generation models, \our is several times faster than sparse attention alone. Furthermore, we implement efficient forward and backward passes for \our. With a few steps of fine-tuning, \our significantly reduces the computation complexity and latency of attention while preserving the quality of the generation results. 

\textbf{Result.} \our reduces attention computation by $\mathbf{95\%}$ without degrading video generation quality, even at a moderate sequence length of 30K, which is the sequence length in Wan2.1-1.3B. In addition, our implementation achieves a $\mathbf{13.7}\times$ speedup in the attention kernel and a $\mathbf{2.2}\times$ end-to-end acceleration for video generation, where the attention time becomes almost negligible. \our consistently surpasses baselines in both generation quality and efficiency.

\section{Preliminary}  \label{sec:preliminary}
\subsection{Block Sparse Attention}
Given queries, keys, and values $Q,K,V \in \mathbb{R}^{N\times d}$, the standard attention computes the score matrix $S = QK^\top/\sqrt{d}$ and the attention weights $P = \mathrm{Softmax}(S)$ to obtain the output $O = PV$. This is inefficient for large $N$ as it requires $\mathcal O(N^2d)$ operations. The idea of sparse attention is to reduce computation by applying a mask $M \in \{0,1\}^{N\times N}$ to the attention weights: $P \gets P \odot M$, where $\odot$ is the element-wise product. A common strategy is to choose a threshold $\tau$ and set $M_{ij}=1$ if $P_{ij}>\tau$. For entries with $M_{ij}=0$, the multiplications $Q_iK_j^\top$ and $P_{ij}V_j$ can be skipped, where $Q_i=Q[i,:], K_j=K[j,:], V_j=V[j,:]$.

However, element-wise sparse attention is inefficient on modern GPUs. Practical implementations such as FlashAttention~\citep{dao2023flashattention} operate at the block level. Specifically, the sparse FlashAttention first partitions $Q, K, V, S, P, M$ into blocks $\{\vQ_i\}, \{\vK_j\}, \{\vV_j\}, \{\vS_{ij}\}, \{\vP_{ij}\}, \{\vM_{ij}\}$, where $\vQ_i \in \mathbb{R}^{b_q \times d}$, $\vK_j,\vV_j \in \mathbb{R}^{b_{kv} \times d}$, and $\vS_{ij}, \vP_{ij}, \vM_{ij} \in \mathbb{R}^{b_q \times b_{kv}}$. Each block mask $\vM_{ij}$ is fully filled with either $0$ or $1$, and we skip the computations of $\vQ_i\vK_j^\top$ and $\vP_{ij}\vV_j$ if $\vM_{ij}[:,:]=0$.

\subsection{Linear Attention} \label{sec:linear_attn}
Linear attention methods reduce the complexity of standard attention from $\mathcal O(N^2d)$ to $\mathcal O(N d^2)$. A key idea is to decouple the softmax operation by introducing a feature map $\phi(\cdot)$ applied to $Q$ and $K$. Specifically, it replaces the attention weights in standard attention with $\frac{\phi(Q)\phi(K)^\top}{{\rm rowsum}(\phi(Q)\phi(K)^\top}$. This reformulation enables reordering of the matrix multiplications: instead of explicitly computing the attention weights, it first computes $\phi(K)^\top V$, and then applies this intermediate result to $\phi(Q)$: 
\[
H = \phi(K)^\top V, \quad Z = {\rm rowsum}(\phi(K)^\top)\in\mathbb R^{d\times 1}, \quad O = {\phi(Q)H\over \phi(Q) Z}.
\]
The mapping $\phi(\cdot)$ is usually an activation function (e.g., $\mathrm{ELU}+1$ or $\mathrm{ReLU}$~\citep{clevert2015elu,glorot2011relu}). This formulation avoids explicitly constructing the $N \times N$ matrices $S, P$ and achieves linear computational complexity.

\section{Motivation and Anlysis}
\begin{figure}[h!]
    \centering
    \vspace{-.6em}
    \includegraphics[width=0.93\textwidth]{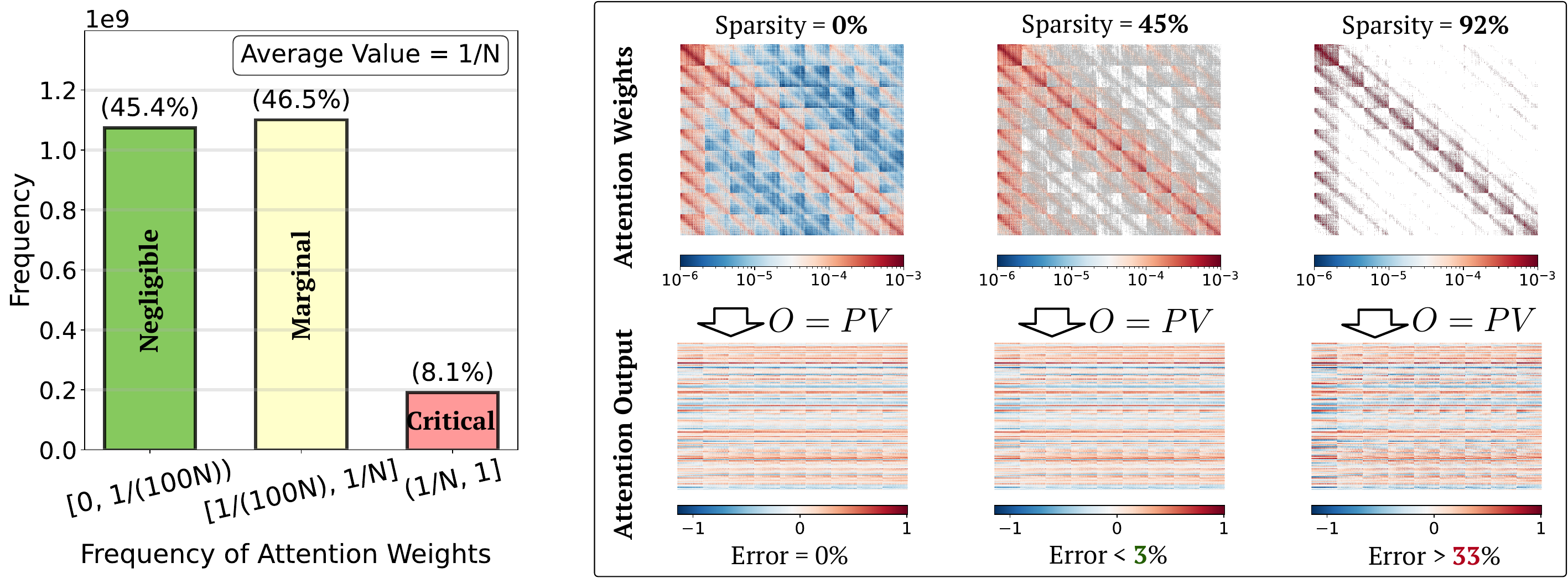}
    \vspace{-.75em}
    \caption{The left figure shows a typical distribution of attention weights sampled from the Wan2.1 model. The right figure shows the accuracy of sparse attention with different sparsity.} \vspace{-.6em}
    \label{fig:p_distribution}
\end{figure}
\subsection{Motivation of \our}
Due to the softmax operator, the attention weights $P$ lie in $[0,1]$ with each row summing to 1. Furthermore, because of the exponential scaling in softmax, only a small fraction of entries in $P$ are relatively large, while the vast majority are close to zero. Figure~\ref{fig:p_distribution} (left) shows the typical distribution of attention weights $P$ sampled from the Wan2.1 model~\citep{wan2025}. We highlight two key observations: (1) Only about $8.1\%$ of the weights are larger than the average value $1/N$. (2) A considerable proportion of weights are extremely small. In our case, approximately $45\%$ fall below $1/(100N)$.
As shown in Figure~\ref{fig:p_distribution} (right), skipping these smallest 45\% of weights in sparse attention (i.e., setting the corresponding entries in $M$ to $0$) introduces a relative L1 error of less than $3\%$ compared to the full attention output. In contrast, retaining only the largest $8.1\%$ of weights (sparsity $=92\%$) leads to a sharp increase in error, reaching about 33\%. This explains why existing sparse attention methods struggle to achieve a sparsity beyond 90\%. 

The intermediate values between $1/(100N)$ and $1/N$ (the yellow column in Figure~\ref{fig:p_distribution}) present a dilemma: omitting them introduces significant accuracy loss, yet computing them with full attention causes a great decrease in sparsity. Fortunately, these values are far less critical than the largest ones.
This finding motivates us to categorize the attention weights into three types: \emph{critical}, \emph{marginal}, and \emph{negligible}. For \emph{critical} weights, we use sparse FlashAttention to compute the output as they dominate the attention distribution; For \emph{negligible} weights, we skip the computation; For \emph{marginal} weights, we employ a linear attention method to reduce the computational complexity to $\mathcal{O}(N d^2)$ and enhance the performance of sparse attention.

\begin{figure}[h!]
    \centering
    \vspace{-.5em}
    \includegraphics[width=0.936\textwidth]{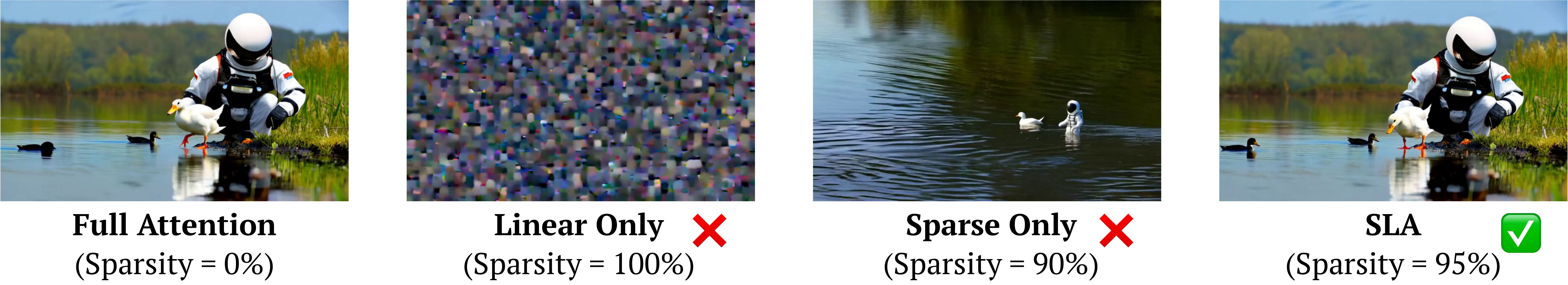}
    \vspace{-.75em}
    \caption{Video generation examples on Wan2.1 fine-tuned with full attention, linear attention, sparse attention, and \our. \our could achieve a high sparsity of 95\% and lossless video quality.}
    \vspace{-.5em}
    \label{fig:motivate_example}
\end{figure}

\uline{Empirical results.} In Figure~\ref{fig:motivate_example}, we present some videos generated by Wan2.1 fine-tuned with different attention methods: using only linear attention, sparse attention with 90\% sparsity, and \our with 95\% sparsity. Note that the computational complexity of \our at 95\% sparsity is nearly half that of 90\% sparse attention, since the cost of linear attention is almost negligible. For example, in the Wan2.1 model, linear attention accounts for less than 0.5\% of the cost of full attention. These empirical results show that \our significantly outperforms the other two methods in video quality.

\subsection{Separating Attention Weights: Sparse Few, Low-Rank Many} \label{sec:low_rank_approximation}

\newtheorem*{observation}{Observation}

\begin{figure}[h!]
    \centering
    \vspace{-.5em}
    \includegraphics[width=0.97\textwidth]{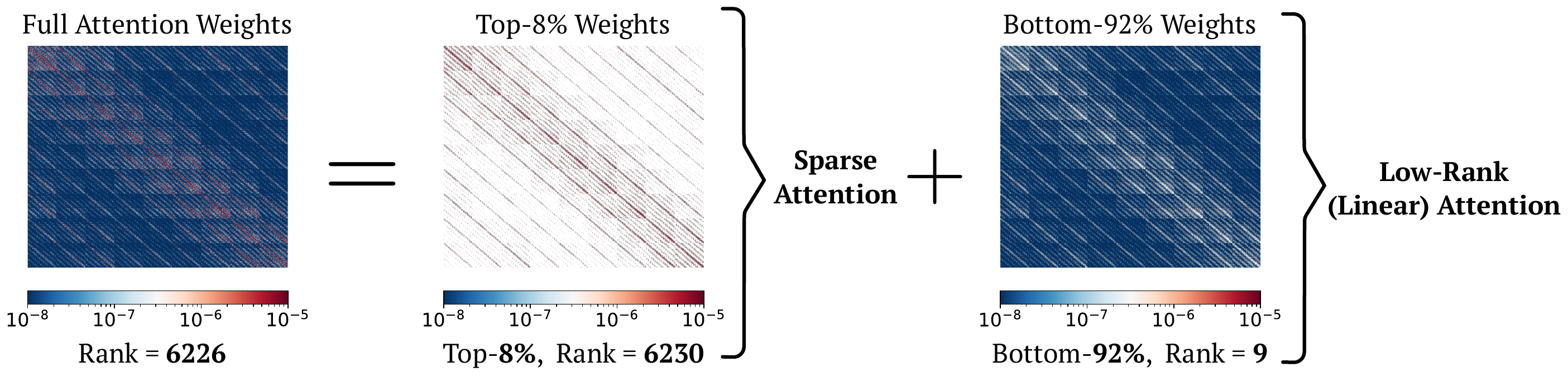}
    \vspace{-.81em}
    \caption{Decomposition of attention weights. We sample attention weights from the Wan2.1 model: the left figure shows the full weights, the middle the top 8\%, and the right the bottom 92\%.}
    \vspace{-.5em}
    \label{fig:analysis}
\end{figure}
\begin{observation}
As shown in Figure~\ref{fig:analysis}, full attention weights can be decoupled into two parts: (1) a small subset ($<10\%$) with rank comparable to full attention, and (2) a large subset ($>90\%$) with very low rank. Since the methods for accelerating attention focus mainly on sparsity or low-rank structure, this suggests a \textbf{natural and elegant strategy}: apply sparse attention to the first part and low-rank approximation to the second.
\end{observation}
Previous failures of linear attention are largely due to the high rank of full attention weights~\citep{fan2025breaking}, while linear attention is restricted to a rank at most $d$. Figure~\ref{fig:analysis} (left) illustrates this with a typical example using the notion of stable rank~\citep{rudelson2006samplinglargematricesapproach}. We observe that after removing the top values in the attention weights $P$, the remaining matrix becomes extremely low-rank. 
This motivates the decomposition of $P$ using the sparse mask $M$:
\begin{align} \label{equ:observation}
   P = \underbrace{P\odot M}_{\text{sparse component}} + \underbrace{P\odot (1-M)}_{\text{low-rank component}}. 
\end{align}

Since linear attention is essentially a low-rank version of attention, we are provided with a possibility to replace the low-rank component $P\odot (1-M)$ with linear attention.

\section{\our} \label{sec:sla_description}
\our effectively integrates sparse and linear attention within a unified framework, allowing them to complement each other. In particular, we fuse both attention into a single efficient GPU kernel. In this section, we introduce the sparse and linear attention components of \our. 

\our first predicts a compressed attention weights matrix $P_c \in \mathbb{R}^{N/b_q \times N/b_{kv}}$:
\begin{equation}
    P_c = \mathrm{Softmax} ( \mathrm{pool}(Q) \mathrm{pool}(K)^\top / \sqrt{d} ).
\end{equation}
where $\mathrm{pool(\cdot)}$ is a mean pooling operator along the token dimension. For each element of $P_c$, we classify it into three types and record the results in a compressed mask $M_c \in \mathbb{R}^{N/b_q \times N/b_{kv}}$. Specifically, the top $k_h\%$ positions are marked as critical (labeled $1$), the bottom $k_l\%$ positions as negligible (labeled $-1$), and the remaining positions as marginal (labeled $0$). Formally,
\begin{equation}
M_c[i,j] = \{1 \;(\text{top }k_h\%),\;
~~-1 \;(\text{bottom }k_l\%),\;
~~0 \;(\text{otherwise})\}.
\end{equation}

We apply different methods according to $M_c$.

\subsection{Sparse Attention in \our}
Guided by the mask $M_c$, sparse FlashAttention is used to compute the sparse attention output. For each $Q$ block $\vQ_i$, we iterate over all $K,V$ blocks $\vK_j, \vV_j$ with $j=0,\dots,N/b_{kv}$. Whenever $M_c[i,j]=1$, we perform:
\begin{align}
\vS_{ij} = \vQ_i \vK_j^\top/\sqrt{d}, ~~~~\vP_{ij} &= \mathrm{OnlineSoftmax}(\vS_{ij}), ~~~~ \vO_i^s = \vO_i^s + \vP_{ij}\vV_j.
\end{align}
Here, $\mathrm{OnlineSoftmax}(\cdot)$ operator~\citep{milakov2018online} computes the softmax of a matrix in a block-wise manner (see lines 10-11 of Algorithm~\ref{alg:fwd} for implementation). The initial value of each $\vO_i^s$ is set to zero. Algorithm~\ref{alg:fwd} describes the forward computation of the sparse attention component, and we denote the final output of the sparse attention component $O^s$.

\subsection{Linear Attention in \our}
Inspired by the idea of low-rank approximation, we replace the low-rank component $P\odot (1-M)$ in Equation~\ref{equ:observation} with linear attention introduced in Section~\ref{sec:linear_attn} as
$$
\frac{\phi(Q)\phi(K)^\top} {{\rm rowsum}(\phi(Q)\phi(K)^\top)} \odot (1-M).
$$
Specifically, the entries of $0$ in $M_c$ determine the blocks processed by linear attention. For each query block $\vQ_i$, we compute the corresponding linear attention output:
\begin{align}
    \vH_i = \sum_{j : M_c[i,j]=0} \phi(\vK_j)^\top \vV_j,
    ~~~~\vZ_i = \sum_{j : M_c[i,j]=0} \mathrm{rowsum}(\phi(\vK_j)^\top),
    ~~~~\vO_i^l = \frac{\phi(\vQ_i)\vH_i}{\phi(\vQ_i)\vZ_i}.
\end{align}
Here, as mentioned in Section~\ref{sec:linear_attn}, $\phi(\cdot)$ denotes the activation function, and $\vH_i\in\mathbb R^{d\times d},\vZ_i\in\mathbb R^{d\times 1}$ are intermediate results similar to $H$ and $Z$.
Algorithm~\ref{alg:fwd} describes the forward pass of the linear attention component, and the final output of this component is denoted as $O^l$.

Finally, the overall attention output of \our is defined as:
\begin{align}
O = O^s + \mathrm{Proj}(O^l).
\end{align}
where $\mathrm{Proj}$ is a learnable linear transformation $\mathbb{R}^d \to \mathbb{R}^d$. Applying this projection to $O^l$ helps reduce the distribution mismatch between softmax and linear attention. Its computational cost is $\mathcal O(Nd^2)$, the same as computing $O^l$ and negligible compared with the $\mathcal O(N^2d)$ cost of full attention.

\textbf{Insight.} Linear attention in \our does not approximate the output corresponding to marginal attention weights, but serves as a learnable compensation that enhances the effectiveness of sparse attention. This is because linear attention alone struggles to approximate the output of full attention~\citep{choromanski2020rethinking, zhen2022cosformer}. Therefore, we need to fine-tune the parameters of the target model, enabling it to adapt to the use of linear attention.

\begin{figure}[h!]
    \centering
    \vspace{-.1em}
    \includegraphics[width=0.999\textwidth]{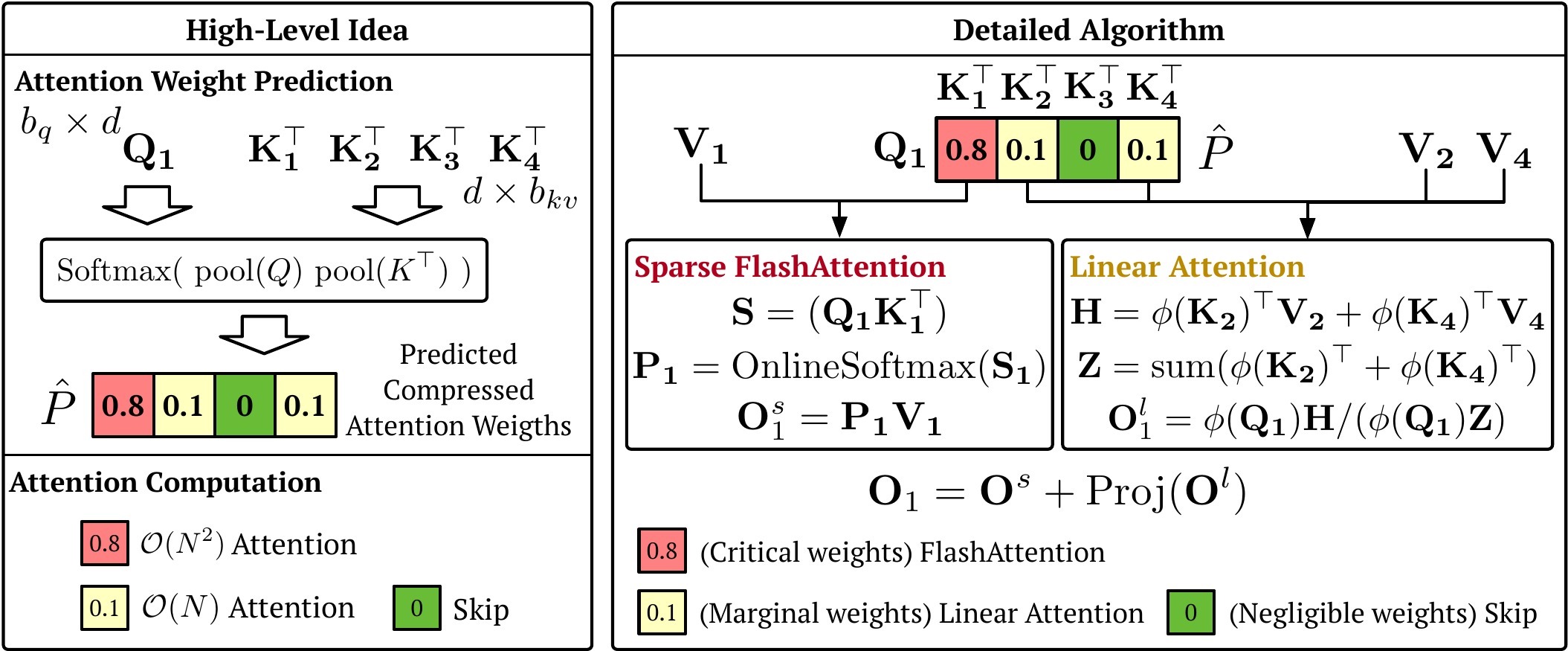}
    \vspace{-.75em}
    \caption{Overview of \our. The left figure illustrates the high-level idea: attention weights are classified into three categories and assigned to computations of different complexity. The right figure shows the detailed forward algorithm of \our using the predicted compressed attention weights.}
    \vspace{-.25em}
    \label{fig:overview}
\end{figure}

\section{Fine-Tuning using \our}
To apply \our to a diffusion model, we can simply replace the original attention with \our and fine-tune the model for a few steps on a dataset consistent with the pretraining data. 
In this section, we describe the forward and backward passes of \our. Moreover, we detail some additional efficiency optimization for \our in Appendix~\ref{sec:extra_algo}.

\begin{algorithm}[h!]
    \small
    \caption{Forward pass of \our.}
    \label{alg:fwd} 
    \begin{algorithmic}[1]
    \STATE {\bf Input:} {Matrices $Q, K, V, Q^\phi, K^\phi \in \mathbb{R}^{N \times d}$, block sizes $b_q, b_{kv}$, hyper-parameters $k_h,k_l$}.
    \STATE {Divide $Q, Q^\phi$ to $T_m = N / b_q$ blocks $\{\vQ_i\}$ and $\{\vQ^\phi_i\}$} ;
    \STATE {Divide $K, V, K^\phi$ to $T_n = N / b_{kv}$ blocks $\{\vK_i\}$, $\{\vV_i\}$ and $\{\vK_i^\phi\}$} ;
    \STATE {$h=\{h_j\}=\{(\vK_j^\phi)^\top \vV_j\}$ ; $z=\{z_j\}=\{{\rm rowsum}((\vK_j^\phi)^\top)$\}} ; ~~~ \annotate{// Precompute for linear attention}
    \STATE {$P_c = {\rm Softmax}({\rm pool}(Q){\rm pool}(K)^\top / \sqrt{d})$ ; $ ~~~~~ $ Initialize $M_c=0$} ;
    \STATE {$M_c[i,j]=1$ if $P_c[i,j]\in{\tt TopK}(P_c[i,:], k_h)$ ; $M_c[i,j]=0$ if $P_c[i,j]\in{\tt BottomK}(P_c[i,:], k_l)$} ;
    \FOR {$i=1$ {\bf to} $T_m$}
        \FOR {$j=1$ {\bf to} $T_n$} 
            \IF {$M_c[i,j]=1$}
                \STATE {$\vS_{ij} = \vQ_i \vK_j^\top / \sqrt{d}$ ; ~~~~~ $m_{ij} = {\rm max}(m_{i, j-1}, {\rm rowmax}(\vS_{ij}))$} ; {$\vP_{ij}=\exp(\vS_{ij}-m_{ij})$} ;
                \STATE {$l_{ij}=e^{m_{i,j-1}-m_{ij}} l_{i,j-1} + {\rm rowsum}(\vP_{ij})$ ; ~~~~~ $\vO_{ij}^s = {\rm diag}(e^{m_{i,j-1}-m_{ij}}) \vO_{i,j-1}^s + \vP_{ij} \vV_j$} ;
            \ELSIF {$M_c[i,j]=0$}
                \STATE $\vH_i \gets \vH_i + h_j ; ~~~~~ \vZ_i \gets \vZ_i + z_j$ ;
            \ENDIF
        \ENDFOR
        \STATE $\vO_i^s={\rm diag}(l_i^{T_n})^{-1}\vO_{i,T_n}^s ; ~~~ \vO_i^l = \vQ_i^\phi \vH_i / (\vQ_i^\phi \vZ_i) ; ~~~ \mathbf{L}_i = m_{i, T_n} + \mathrm{log}(l_{i, T_n})$ ;
    \ENDFOR
    \STATE \textbf{return} $O^s=\{\vO^s_i\}$,~~~$O^l=\{\vO^l_i\}$ ;
    \end{algorithmic}
\end{algorithm}

\subsection{Forward Pass}

The formulation of the forward computation was introduced in Section~\ref{sec:sla_description}. The complete algorithm of the forward pass of \our is presented in Algorithm~\ref{alg:fwd}. It's worth noting that we precompute $h_j=\phi(\vK_j)^\top\vV_j$ and $z_j={\rm rowsum}(\phi(\vK_j)^\top)$ for each pair $(K_j, V_j)$ (Line 4 in Algorithm~\ref{alg:fwd}). This design ensures that, when $M_c[i,j] = 0$, the corresponding operation only involves a single matrix addition (Line 13 in Algorithm~\ref{alg:fwd}), thereby improving efficiency. To simplify the notation, we denote $Q^\phi = \phi(Q)$ and $K^\phi = \phi(K)$ in the following.

\begin{algorithm}[h!]
    \small
    \caption{Backward pass of \our.}
    \label{alg:bwd} 
    \begin{algorithmic}[1]
    \STATE {\bf Input:} {$Q, K, V, Q^\phi, K^\phi, M_c, \{\mathbf L_i\}, \{\vH_i\}, \{\vZ_i\}, O^s, O^l $ from the forward, $dO^s,dO^l \in \mathbb R^{N\times d}$}.
    \STATE {$D^s={\rm rowsum}(dO^s\odot O^s)$, $D^l={\rm rowsum}(dO^l\odot O^l)$, divide $D^s, D^l$ into $T_m$ blocks $\{\vD_i^s\},\{\vD_i^l\}$} ;
    \FOR {$i=1$ {\bf to} $T_m$}
        \STATE {$\vdH_i=(\vQ_i^\phi/(\vQ^\phi_i \vZ_i))^\top \vdO^l_i$; ~ $\vdZ_i=-(\vQ_i^\phi/(\vQ_i^\phi \vZ_i))^\top D_i^l$} ;
        \STATE {$\vdQ^\phi_i=(\vdO^l_i (\vH_i)^\top - \vD_i^l\vZ_i^\top) / (\vQ^\phi_i \vZ_i)$} ;
    \ENDFOR
    \FOR {$j=1$ {\bf to} $T_n$}
        \STATE {Initialize $\vdH=0,\vdZ=0$} ;
        \FOR {$i=1$ {\bf to} $T_m$}
            \IF {$M_c[i,j]=1$}
                \STATE {$\vS_{ij} = \vQ_i \vK_j^\top / \sqrt{d}$} ;~~ {$\vP_{ij}=\exp(\vS_{ij}-\mathbf L_i)$ ; ~~ $\vdV_j\gets\vdV_j + \vP_{ij}^\top\vdO_i^s$} ;~~ $\vdP_{ij}=\vdO^s_{ij}\vV_j^\top$ ;
                \STATE {$\vdS_{ij}=\vP_{ij}\odot(\vdP_{ij}-\vD_i^s)$} ;~~~ {$\vdQ_i\gets\vdQ_i + \vdS_{ij}\vK_j$ ; ~~~~~ $\vdK_j\gets\vdK_j + \vdS_{ij}^\top\vQ_i$} ;
            \ELSIF {$M_c[i,j]=0$}
                \STATE $\vdH \gets \vdH + \vdH_i ; ~~~~~ \vdZ \gets \vdZ + \vdZ_i$ ;
            \ENDIF
        \ENDFOR
        \STATE {$\vdK^\phi_j=\vV_j(\vdH)^\top+(\vdZ)^\top ; ~~~~~ \vdV_j=\vK^\phi_j\vdH$} ;
    \ENDFOR
    \STATE \textbf{return} $dQ=\{\vdQ_i\}$,~~~$dK=\{\vdK_i\}$,~~~$dV=\{\vdV_i\}$,~~~$dQ^\phi=\{\vdQ^\phi_i\}$,~~~$dK^\phi=\{\vdK^\phi_i\}$ ;
    \end{algorithmic}
\end{algorithm}

\vspace{-1em}
\subsection{Backward Pass}

The backward pass computes gradients for both the sparse and linear components, which are also fused into a single GPU kernel for efficiency.  

\textbf{Gradient notation.} The prefix $d$ or $\bf d$ is used to denote gradients, e.g., $dO^s,dO^l$ are the gradients of $O^s,O^l$ with respect to some loss function $\ell$, respectively.

\textbf{Sparse attention gradients.} The output gradient $dO^s$ is backpropagated to compute $dQ$, $dK$, and $dV$, following the same derivation as in FlashAttention~\citep{dao2023flashattention}. Given $dO^s$, the backward pass is carried out as follows:
\begin{gather}
    \begin{split}
        \vdP_{ij}=\vdO_{ij}^s\vV_j^\top,\quad\vD_i^s={\rm rowsum}(\vdO_i^s\odot\vO_i^s),&\quad\vdS_{ij}=\vP_{ij}\odot (\vdP_{ij} -\vD_i^s),\\
        \quad\vdQ_i=\vdS_{ij}\vK_j,\quad\vdK_j=\vdS_{ij}^\top\vQ_i,&\quad\vdV_j=\vP_{ij}^\top\vdO_{i}^s.
    \end{split}
\end{gather}
Here, we consider $\vD_i^s \in\mathbb R^{b_q\times 1}$ as a column vector.

\textbf{Linear attention gradients.} The gradient $dO^l$ yields $dQ^\phi, dK^\phi, dV$ through the chain rule:
\begin{gather}
    \begin{split}
    \vdH_i=\left(\vQ_i^\phi\over\vQ_i^\phi\vZ_i\right)^\top\vdO^l_i,\quad\vD^l_i={\rm rowsum}(\vdO_i^l\odot\vO_i^l),\quad\vdZ_i=-\left(\vQ_i^\phi\over\vQ_i^\phi\vZ_i\right)^\top\vD_i^l\\
        \vdQ_i^\phi={(\vdO^l_i(\vH_i)^\top - \vD_i^l \vZ_i^\top)\over \vQ_i^\phi \vZ_i},\quad\vdK^\phi_j=\vV_j(\vdH_i)^\top+(\vdZ_i)^\top,\quad \vdV_j=\vK^\phi_j\vdH_i
    \end{split}
\end{gather}
Here, $\vdK_j^\phi$ and $\vdV_j$ are obtained by aggregating $\vdH_i$ and $\vdZ_i$. Similar to the forward pass, each $\vdH_i$ and $\vdZ_i$ is precomputed so that the remaining computation reduces to simple matrix additions. The detailed algorithm is provided in Algorithm~\ref{alg:bwd}.

\section{Experiment}  
\subsection{Setup} \label{sec:exp_setup} 
 \textbf{Model and Datasets.} We use the Wan2.1-1.3B model~\citep{wan2025} for video generation experiments in the main text and LightningDiT~\citep{yao2025vavae} for image generation experiments in the Appendix~\ref{sec:image_exp}.
For video experiments, we use a private dataset collected from websites such as Pexels~\citep{pexels} and Common Crawl~\citep{commoncrawl}, consisting of 20{,}000 5-second videos at 480p resolution for fine-tuning.  
For image experiments, following LightningDiT~\citep{yao2025vavae}, we use the ImageNet~\citep{deng2009imagenet} dataset at a resolution of $512\times512$.

\textbf{Baselines.} We compare \our with state-of-the-art sparse attention methods applicable to diffusion models, including (1) \texttt{VSA}~\citep{zhang2025vsa}, (2) \texttt{VMoBa}~\citep{wu2025vmoba}, and (3) the training-free SparseAttn~\citep{zhang2025spargeattn} (\texttt{Sparge-F}) and (4) a trainable implementation of SpargeAttn (\texttt{Sparge-T}). 
For \texttt{VSA} and \texttt{VMoBa}, we use their official implementations, while for \texttt{Sparse-T}, we implement the method ourselves because there is no official implementation. 
In addition, we design several baselines for ablation studies:  
(5) \texttt{Linear Only}, which applies only linear attention;  
(6) \texttt{Sparse Only}, which applies only the sparse attention component of \our; and (7) \texttt{L+S}, which directly sums the attention outputs of the \texttt{Linear Only} and \texttt{Sparse Only}. 

\textbf{Metrics.} For video quality, following ~\citet{zhang2024evaluationagent,yang2024cogvideox}, we use four evaluation dimensions of VBench~\citep{zhang2024evaluationagent}: Imaging Quality ({\texttt{IQ}}), Overall Consistency ({\texttt{OC}}), Aesthetic Quality ({\texttt{AQ}}), Subject Consistency ({\texttt{SC}}). We also use the Vision Reward ({\texttt{VR}})~\citep{xu2024visionreward} for human preference evaluation, Aesthetic Video Quality ({\texttt{VA}}), and Techniqual Video Quality ({\texttt{VT}})~\citep{liu2023evalcrafter}. For image quality, following ~\citet{yao2025vavae}, we use {\texttt{FID}}. For attention computation complexity, we use {\texttt{FLOPs}} (floating point of operations). For attention efficiency, we use {\texttt{FLOPS}} (floating-point operations per second) for attention kernel efficiency. Specifically, \texttt{FLOPS} here is $\mathcal{O}($full attention$)/t$, where $\mathcal{O}(\cdot)$ denotes the operation count and $t$ the attention latency. We use seconds for end-to-end generation latency.

\textbf{Hyper-parameters.} We use a training batch size of 64 and fine-tune the Wan2.1 model for 2000 steps. For the activation function $\phi$, we use softmax according to our ablation experiments. $k_h$\% is 5\% and $k_l$\% is 10\%. For block size, we use $b_q=b_{kv}=64$. The hyper-parameters for image generation tasks are detailed in Appendix~\ref{sec:image_exp}.

\subsection{Effectiveness} \label{sec:video_effectiveness}
Table~\ref{table:exp_effectiveness} compares the video generation quality and efficiency of \our with baseline methods on Wan2.1-1.3B, fine-tuned separately with \our, Full Attention, and each baseline. \our delivers about a $\bf 19.3\times$ efficiency gain while maintaining video quality comparable to Full Attention. Moreover, compared with the baselines, \our consistently achieves higher quality even under greater sparsity. For example, 95\% (1-\textbf{5}\%) sparsity in \our is actually about $\bf 3\times$ more efficient than 85\% (1-\textbf{15}\%) while still producing better video quality.

\begin{table}[h]
    \centering
    \caption{Quality and efficiency comparison of \our and other baseline methods. }
    \small
    \label{table:exp_effectiveness}
    \setlength\tabcolsep{6.9pt}
    \begin{tabular}{c|ccccccc|cc}
    \toprule
    \multirow{2}{*}{\textbf{Method}} 
    & \multicolumn{7}{c|}{\textbf{Quality}} & \multicolumn{2}{c}{\textbf{Efficiency}} \\
    \cmidrule(lr){2-8} \cmidrule(lr){9-10} & \texttt{VA} $\uparrow$ & \texttt{VT} $\uparrow$ & \texttt{IQ} $\uparrow$ & \texttt{OC} $\uparrow$ & \texttt{AQ} $\uparrow$ & \texttt{SC} $\uparrow$ & \texttt{VR} $\uparrow$ & \texttt{FLOPs} $\downarrow$ & Sparsity $\uparrow$ \\
    \midrule
    Full Attention        & 76.78 & 82.88 & 62.5 & 23.3 & 56.1 & 93.0 &  0.059 & 52.75T & 0\% \\
    \midrule
    \texttt{Sparge-F}     & 0.002 & 0.026 & 26.0 &  4.6 & 35.7 & 85.1 & -0.216 &  7.91T & 85\% \\
    \texttt{Sparge-T}     & 73.83 & 77.87 & 61.9 & 22.7 & 55.4 & 93.1 &  0.014 &  7.38T & 84\% \\
    \texttt{VMoBa}        & 32.33 & 35.79 & 58.0 & 18.8 & 46.2 & 89.9 & -0.175 &  7.91T & 85\% \\
    \texttt{VSA}          & 55.37 & 64.61 & 60.6 & 22.4 & 51.9 & 83.6 & -0.069 &  5.92T & 89\% \\
    \rowcolor{lightblue}
    \our & \textbf{\textcolor{darkgreen}{76.96}} & \textbf{\textcolor{darkgreen}{83.92}} & \textbf{\textcolor{darkgreen}{62.2}} & \textbf{\textcolor{darkgreen}{23.6}} & \textbf{\textcolor{darkgreen}{55.9}} & 93.1 & \textbf{\textcolor{darkgreen}{0.048}} & \textbf{\textcolor{darkgreen}{2.74T}} & \textbf{\textcolor{darkgreen}{95\%}} \\
    \bottomrule
    \end{tabular}
    \vspace{-1.75em}
\end{table}

\begin{table}[h]
    \centering
    \caption{Ablation results for \our.}
    \small
    \label{table:exp_ablation}
    \setlength\tabcolsep{6.47pt}
    \begin{tabular}{c|ccccccc|cc}
    \toprule
    \multirow{2}{*}{\textbf{Method}} 
    & \multicolumn{7}{c|}{\textbf{Quality}} & \multicolumn{2}{c}{\textbf{Efficiency}} \\
    \cmidrule(lr){2-8} \cmidrule(lr){9-10} & \texttt{VA} $\uparrow$ & \texttt{VT} $\uparrow$ & \texttt{IQ} $\uparrow$ & \texttt{OC} $\uparrow$ & \texttt{AQ} $\uparrow$ & \texttt{SC} $\uparrow$ & \texttt{VR} $\uparrow$ & \texttt{FLOPs} $\downarrow$ & Sparsity $\uparrow$ \\
    \midrule
    Full Attention       & 76.78 & 82.88 & 62.5 & 23.3 & 56.1 & 93.0 & 0.059  & 52.75T & 0\% \\
    \midrule
    \texttt{Linear Only} & 0.042 & 0.099 & 39.5 &  3.6 & 28.8 & 90.7 & -0.213 &  0.10T & 100\% \\
    \texttt{Sparse Only} & 64.00 & 70.50 & 57.2 & 21.8 & 51.7 & 88.7 & -0.073 &  7.91T & 85\% \\
    \texttt{L+S}         & 29.65 & 41.15 & 58.6 & 18.8 & 45.3 & 87.1 & -0.105 &  5.37T & 90\% \\
    \rowcolor{lightblue}
    \our (softmax)       & \textbf{76.96} & \textbf{83.92} & 62.2 & \textbf{23.6} & \textbf{55.9} & 93.1 &  0.048 &  \textbf{2.73T} & \textbf{95\%} \\
    \our (elu+1)         & 75.50 & 81.01 & 62.8 & 23.5 & 55.3 & 92.9 &  0.034 &  2.74T & 95\% \\
    \our (hedgehog)      & 74.59 & 82.62 & 61.9 & 22.5 & 54.3 & 93.2 &  0.035 &  3.11T & 95\% \\
    \rowcolor{lightblue}
    \our (Top 5\%)       & 76.96 & 83.92 & 62.2 & 23.6 & 55.9 & 93.1 &  0.048 &  2.73T & 95\% \\
    \our (Top 10\%)      & 75.29 & 82.20 & 62.5 & 22.6 & 55.8 & 93.5 &  0.057 &  5.38T & 90\% \\
    \our (Top 20\%)      & 75.81 & 83.82 & 62.7 & 22.4 & 54.5 & 92.6 &  0.059 & 10.65T & 80\% \\
    \bottomrule
    \end{tabular}
    \vspace{-.5em}
\end{table}

\begin{figure}[h!]
    \centering
    \vspace{-.5em}
    \includegraphics[width=1.0\textwidth]{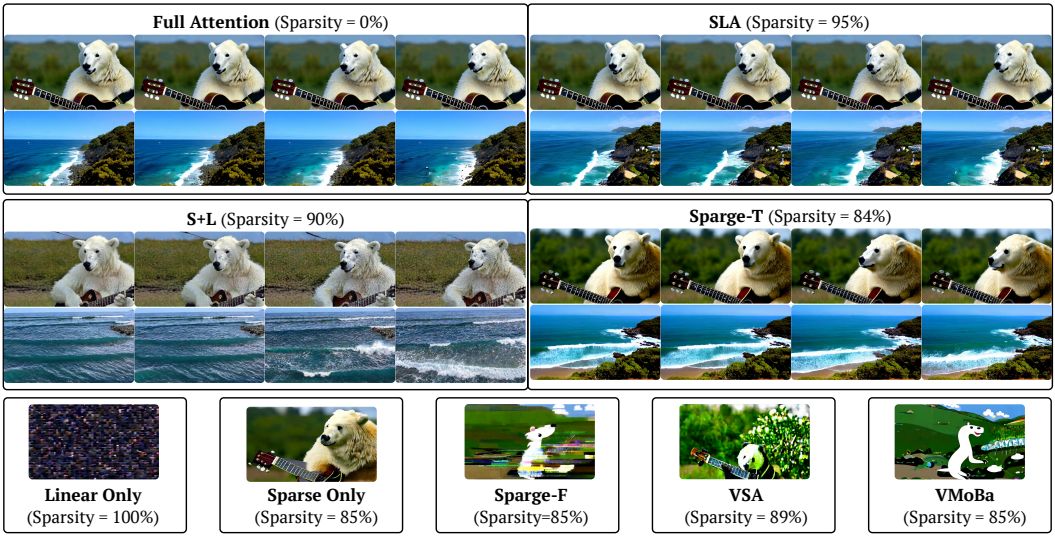}
    \vspace{-1.5em}
    \caption{Video examples using Wan2.1 fine-tuned with \our and baselines. For \texttt{Linear Only}, \texttt{Sparse Only}, \texttt{Sparge-F}, \texttt{VSA}, and \texttt{VMoBa}, only a single frame per prompt is shown, as their video quality is not sufficient. The full visible comparison is in Figure~\ref{fig:video_example_appendix} in Appendix~\ref{appendix:visible_examples}.}
    \vspace{-.72em}
    \label{fig:video_example_main}
\end{figure}

\begin{figure}[h]
    \centering
    \vspace{-.39em}
    \includegraphics[width=1.0\textwidth]{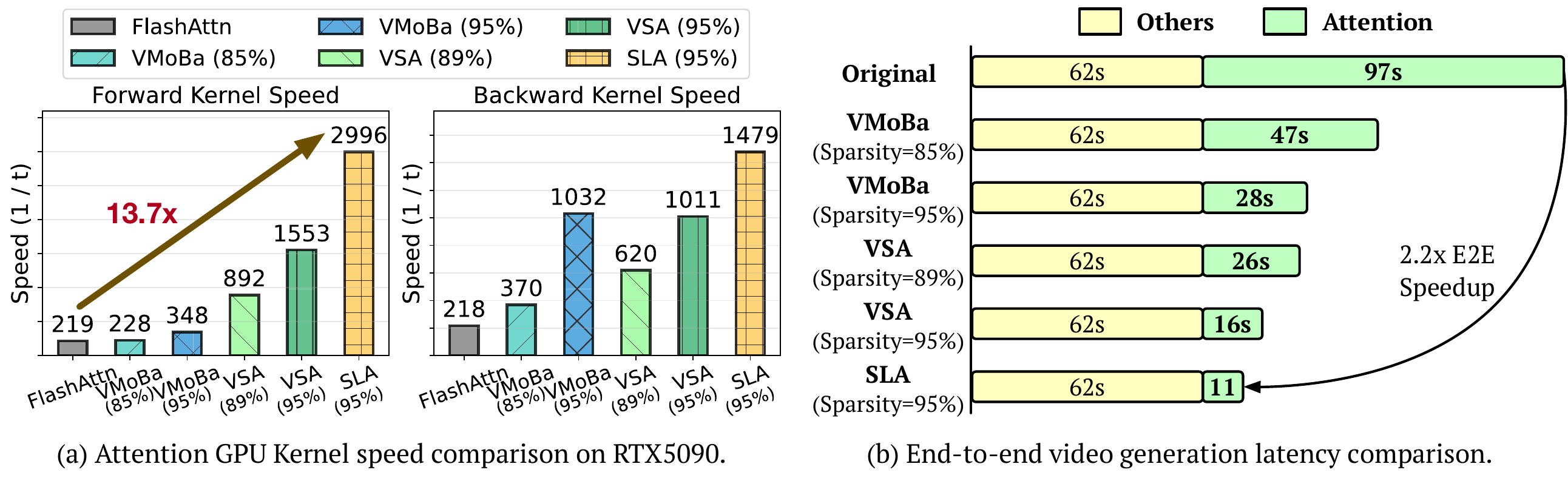}
    \vspace{-2.1em}
    \caption{Attention kernel speed and end-to-end generation latency of \our and baselines on Wan2.1-1.3B with RTX5090. FlashAttn refers to FlashAttn2, the fastest available version on RTX5090.}
    \vspace{-1em}
    \label{fig:exp_acceleration}
\end{figure}

\subsection{Efficiency}
Figure~\ref{fig:exp_acceleration} compares the kernel speed and end-to-end latency of \our on Wan2.1–1.3B with an RTX5090. Note that even \texttt{VSA} in 89\% sparsity and \texttt{VMoBa} in 85\% sparsity, their generation quality is already worse than \our, so higher sparsity settings (e.g., 95\%) are not quality-matched comparisons.
In the forward pass, \our achieves a $\mathbf{13.7}\times$ speedup over FlashAttention2 and is $\mathbf{1.93}\times$ faster than \texttt{VSA} with 95\% sparsity and $\mathbf{3.36}\times$ faster than \texttt{VMoBa} with 95\% sparsity. In the backward pass, it delivers a $\mathbf{6.8}\times$ speedup over FlashAttention2, still outperforming \texttt{VSA} and \texttt{VMoBa}. 
For end-to-end video generation, \our reduces attention latency from 97s to 11s ($\mathbf{8.8}\times$ reduction), resulting in a $\mathbf{2.2}\times$ end-to-end speedup. For fine-tuning overhead, we train Wan2.1-1.3B for only 2,000 steps with a batch size of 64, which is less than 0.1\% of the cost of pretraining (typically $10^5$–$10^6$ steps with a batch size of $10^3$–$10^4$)~\citep{wan2025}.

\subsection{Ablation Study}
\textbf{Fusing sparse and linear attention.} 
To evaluate the effectiveness of SLA in integrating sparse and linear attention, we compare SLA with \texttt{Sparse Only}, \texttt{Linear Only}, and \texttt{S+L} on Wan2.1 in terms of end-to-end generation quality and efficiency. 
As shown in Table~\ref{table:exp_ablation}, \our achieves the best generation quality and is more efficient than \texttt{Sparse Only} and \texttt{S+L}, confirming the effectiveness of our fusion strategy.  

\textbf{Activation function in linear attention.} 
To study the effect of the activation function $\phi$ in the linear attention component of \our, we evaluate softmax, elu+1, and hedgehog. 
Results in Table~\ref{table:exp_ablation} show that softmax generally provides better generation quality and efficiency.  

\textbf{Impact of parameter $k_h$.} 
We vary $k_h$ from 5\% to 20\% and report the results in Table~\ref{table:exp_ablation}. 
We find that $k_h=5\%$ already yields generation quality close to that of full attention. 
Since $k_h=5\%$ saves about half and a quarter of the computation compared with $k_h=10\%$ and $k_h=20\%$, it offers the best trade-off between efficiency and quality.

\subsection{Visible Examples}
Figure~\ref{fig:video_example_main} and Figure~\ref{fig:video_example_appendix} show video examples from Wan2.1-1.3B fine-tuned using \our and baselines. 
\our produces videos comparable to full attention even at \textbf{95}\% sparsity, while other methods exhibit noticeable distortions even at sparsity levels below 90\%.

\vspace{-.5em}
\section{Related Work} \vspace{-.5em}
As sequence lengths in generative models (e.g., language and video) grow, the quadratic cost of attention becomes a key bottleneck~\citep{zhangsurvey}. Many studies aim to improve efficiency in two main directions: sparse and linear attention. Most sparse attention methods~\citep{xiao2023efficient,xiao2024infllm,jiang2024minference,gao2024seerattention,moaattention,xi2025sparse,zhang2025spargeattn,ribar2023sparq,yang2025sparse} speed up inference without training by masking computation at test time. Some~\citep{zhang2025vsa,wu2025vmoba} add sparsity during training, enabling higher sparsity. Linear attention methods~\citep{wang2020linformer,choromanski2020rethinking,katharopoulos2020transformers,qin2024lightning,yang2024gated,sun2023retentive} are mainly studied in language models. For DiT, SANA~\citep{xie2024sana} and Dig~\citep{zhu2025dig} show linear attention works for image generation pre-training, but in video generation, existing methods cannot rely on it alone for lossless quality. Another direction is hardware-efficient attention~\citep{dao2022flashattention,dao2023flashattention,shah2024flashattention,zhang2024sageattention,zhang2024sageattention2,zhang2025sageattention2++}, which optimizes GPU execution through tiling, kernel fusion, and quantization.

\vspace{-.5em}
\section{Conclusion}
\vspace{-.5em}
We propose \our, a trainable attention that unifies sparse and linear attention to accelerate Diffusion Transformers. 
\our assigns computation according to importance: it computes $\mathcal{O}(N^2)$ attention for critical weights, $\mathcal{O}(N)$ attention for marginal weights, and skips negligible computations. 
This design enables substantial reductions in attention cost while preserving effectiveness.  
Experiments show that just a few fine-tuning steps enable \our to accelerate models effectively. Specifically, \our achieves a $\bf 20\times$ reduction in attention computation, along with a $\bf 13.7\times$ GPU kernel speedup and a $\bf 2.2\times$ end-to-end speedup on Wan2.1-1.3B, all without degrading the quality of video generation.



\bibliography{main}
\bibliographystyle{iclr2026_conference}

\newpage

\appendix

\section{Appendix}

\subsection{More Visible Examples}  \label{appendix:visible_examples}

\begin{figure}[h!]
    \centering
    \includegraphics[width=1.0\textwidth]{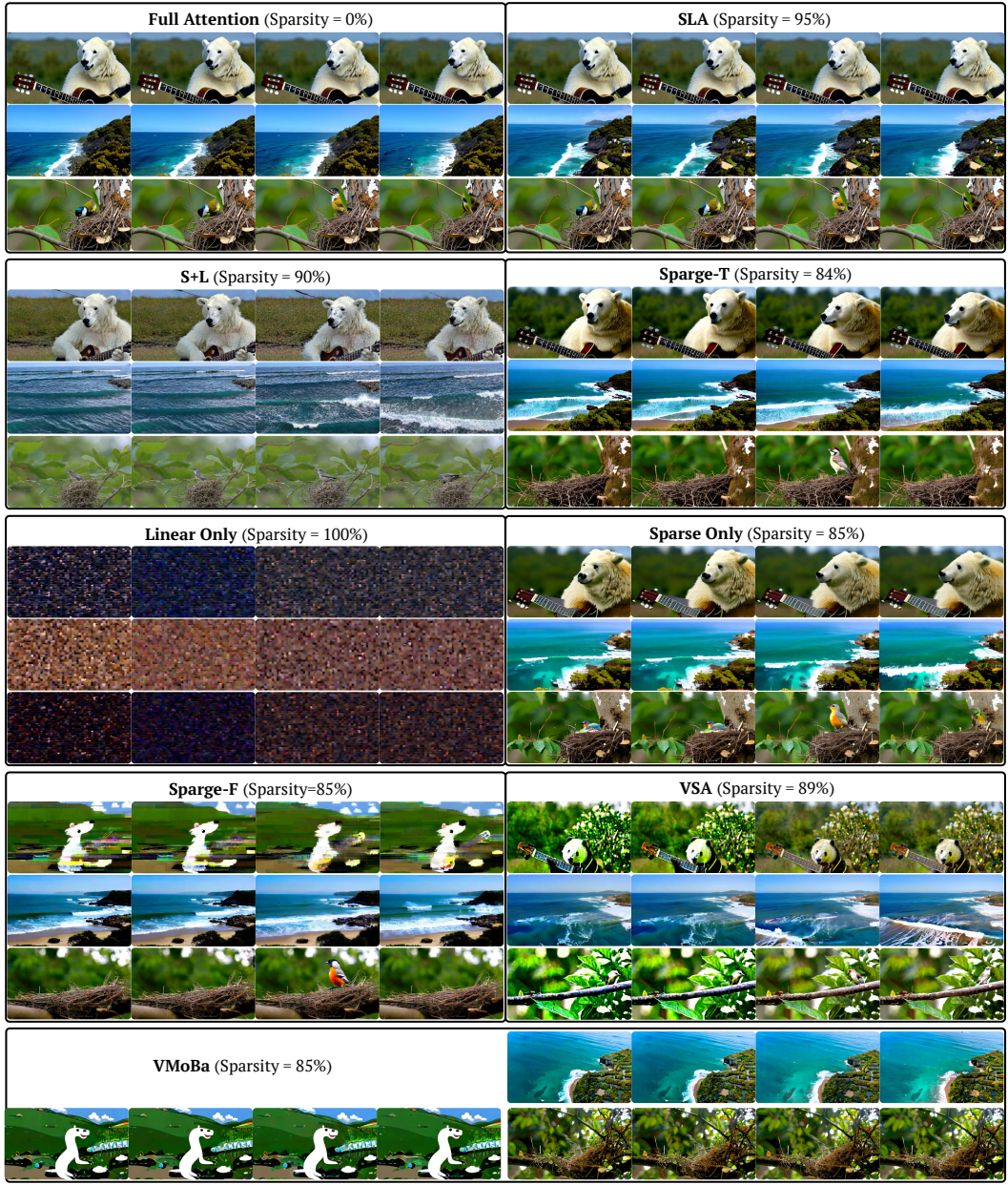}
    \vspace{-1em}
    \caption{Full video examples generated by the Wan2.1 fine-tuned with \our and baseline methods. The first prompt is \textit{``A polar bear is playing guitar"}. The second prompt is \textit{``Pacific coast, carmel by the sea ocean and waves"}. The third prompt is \textit{``a bird building a nest from twigs and leaves"}.}
    \label{fig:video_example_appendix}
\end{figure}

Figure~\ref{fig:video_example_appendix} demonstrates some additional video examples generated by the Wan2.1 model, fine-tuned with \our and other attention methods. We can find that \our consistently achieves higher quality even under bigger sparsity than baselines.

\subsection{Experiments for Image Generation} \label{sec:image_exp}

\begin{table}[!h]
    \centering
    \caption{Quality and efficiency comparison of \our and other baselines on image generation.}
    \label{table:exp_image}
    \setlength\tabcolsep{18pt}
    \begin{tabular}{c|c|cc}
    \toprule
    \multirow{2}{*}{\cellcolor{white}\textbf{Method}}
    & \multicolumn{1}{c|}{\textbf{Quality}} & \multicolumn{2}{c}{\textbf{Efficiency}} \\
    \cmidrule(lr){2-2} \cmidrule(lr){3-4} & \texttt{FID} $\downarrow$ & FLOPs $\downarrow$ & Sparsity $\uparrow$ \\
    \midrule
    Full Attention        &  31.87 & 12.88G & 0\% \\
    \midrule
    \texttt{SpargeAttn-F} & 206.11 &  3.66G & 71.57\% \\
    \texttt{SpargeAttn-T} &  46.05 &  3.16G & 75.45\% \\
    \texttt{VSA(2D)}      &  35.75 &  3.62G & 75.00\% \\
    \texttt{VMoBA(2D)}    &  39.45 &  3.22G & 75.00\% \\
    \rowcolor{lightblue} 
    \our & \textbf{\textcolor{darkgreen}{31.49}} & \textbf{\textcolor{darkgreen}{1.73G}} & \textbf{\textcolor{darkgreen}{87.50\%}} \\
    \rowcolor{lightblue}
    \bottomrule
    \end{tabular}
\end{table}

\textbf{Experimental setup.} As described in Section~\ref{sec:exp_setup}, we evaluate \our and the baselines on a pretraining task of LightningDiT~\citep{yao2025vavae}. Specifically, we use the \texttt{LightningDiT-1p0B/1} model, consisting of 1.03B parameters, trained on the ImageNet~\citep{deng2009imagenet} dataset at a resolution of $512 \times 512$.

\textbf{Hyperparameters.} All hyperparameters follow~\citep{yao2025vavae}, except that we train for $100000$ steps with a batch size of $128$. For \our, we set $\phi$ to softmax and use a block size of $b_q=b_{kv}=64$.

\textbf{Metrics.} Following~\citep{yao2025vavae}, we adopt \texttt{FID} to assess image quality and FLOPs to measure computational complexity.

\textbf{Results.} The results are summarized in Table~\ref{table:exp_image}. At the highest sparsity level, \our outperforms all other baselines and \uline{even surpasses full attention} on the \texttt{FID} metric, confirming the advantage of \our in preserving image quality. This finding is consistent with the video experiments on Wan2.1 reported in Section~\ref{sec:video_effectiveness}.

\subsection{Additional Efficiency Optimization} \label{sec:extra_algo}

Since the efficiency of \our depends heavily on the sparsity pattern, we introduce several complementary optimizations tailored to different sparsity levels. These optimizations lead to substantial gains in computational efficiency:  

\textbf{Lookup table.} When $M_c$ is highly sparse (e.g., sparsity $>90\%$), scanning entire rows or columns to read mask values causes significant memory overhead. To mitigate this, we preprocess the nonzero positions of each row and column and store them in a lookup table. During computation, only the lookup table is accessed, substantially reducing memory traffic.  

\textbf{Pre-aggregation for linear attention.} Although Line 13 in Algorithms~\ref{alg:fwd} and Line 14 in Algorithm~\ref{alg:bwd} require only a single matrix addition, repeatedly performing such additions incurs high overhead when many entries of $M_c$ are $0$ (e.g., $>90\%$). To address this, we precompute the row/column sums $\sum_j h_j$ and $\sum_j z_j$, and then subtract the contributions corresponding to $M_c[i,j]\neq 0$. In this way, $90\%$ of the additions can be replaced by only $10\%$ subtractions.

\textbf{Method of Four Russians.} When the number of blocks with $M_c[i,j] = 0$ is neither very small nor very large (e.g., around $50\%$), we provide an efficient implementation for Line 13 in Algorithms~\ref{alg:fwd} and Line 14 in Algorithm~\ref{alg:bwd}. Specifically, we adopt the Method of Four Russians~\citep{Arlazarov1970TransitiveClosure}. The key idea is to group ${h_j}$ and ${z_j}$ into segments of $g$ consecutive blocks and precompute all $2^g$ possible subset sums within each segment. During the forward pass, any subset of $g$ blocks can then be obtained by a single look-up, rather than summing them on the fly. This scheme allows a theoretical computation reduction by $1/g$.

\section*{Use of Large Language Models}
We used a language model only for polishing English writing, while all ideas, experiments, results, and interpretations are our own.

\end{document}